# A Comparative Analysis of Transfer Learning-based Techniques for the Classification of Melanocytic Nevi

Sanya Sinha[1][0000-0001-5337-635X] and Nilay Gupta[2][0000-0002-6319-8056]

[1] Birla Institute of Technology Mesra, Patna Campus, Patna, India
ssanya0904@gmail.com
[2] Sikkim Manipal Institute of Technology, Majitar, India
ng.nilaygupta04@gmail.com

**Abstract** Skin cancer is a fatal manifestation of cancer. Unrepaired deoxyribonucleic acid (DNA) in skin cells, causes genetic defects in the skin and leads to skin cancer. To deal with lethal mortality rates coupled with skyrocketing costs of medical treatment, early diagnosis is mandatory. To tackle these challenges, researchers have developed a variety of rapid detection tools for skin cancer. Lesion-specific criteria are utilized to distinguish benign skin cancer from malignant melanoma. In this study, a comparative analysis has been performed on five Transfer Learning-based techniques that have the potential to be leveraged for the classification of melanocytic nevi. These techniques are based on deep convolutional neural networks (DCNNs) that have been pre-trained on thousands of open-source images and are used for day-to-day classification tasks in many instances.

## 1 Introduction

Melanoma is an uncommon and fatal variation of skin cancer known to affect nearly 100,000 people annually [1]. According to the American Cancer Society, melanoma accounts for just 1% of all lesions, but it has a higher mortality rate. Melanoma malignancies may only be cured if detected early. They are prone to infect other susceptible cells and spread casualty.

Melanoma can be treated effectively only if it has been diagnosed at an early stage, because it gets increasingly difficult to cure as it progresses. As a result, prognosis and accurate detection are critical for operative treatment [2]. Dermoscopy pictures are used to confirm melanoma. [3] However, dermatologists have a sensitivity rate of less than 80% for the unassisted diagnosis of dermoscopic pictures [4]. Consequently, computer-aided melanoma diagnosis has the potential to increase the sensitivity of melanoma detection.

Nonetheless, the dermoscopy image classification process remains perplexing due to anomalies and noise in nevus shapes caused due to variations in light due to defective imaging. If even a solitary pixel of melanoma was found in non-malignant nevi, the entire nevus was classed as melanoma, even if spectral distributions of lesions overlapped. Thus, an explicit understanding of an aberrant condition of use is required, especially for feature extraction. But, selecting an acceptable feature is considerably time-consuming [5].

Transfer Learning (TL) is a popular approach in deep learning which minimizes the need for feature selection and extraction. Pre-trained models are used as the starting point. This approach effectively minimizes the time and compute resources required to train DCNNs from scratch. TL is a paradigm in which a model generated for one job is reutilized as the wireframe for a model on a diverse use case.

Traditional machine learning employs an isolated training technique in which each model is trained individually for a specific goal with no reliance on prior information. TL, on the other hand, employs information learned from the pre-trained model to complete the job. Consequently, TL models outperform typical machine learning paradigms in terms of performance and effectively reutilize resources and weights from previously trained models. It is faster than training neural networks from the ground up.

## 2   Related Work

To pave the way for automated diagnosis in the melanoma landscape, a number of machine learning-based methods have been developed. Mohd et al. [6] formulated a classification algorithm for classifying four types of skin cancers by using the K-means clustering method. Similarly, Almansour et al. [7] fabricated an end-to-end melanoma classification system using K-means clustering followed by the Support Vector Machine (SVM) algorithm. Abbas et al. [8] developed a novel solution with multi-component patterns and an adaptive boosting multi label learning algorithm (AdaBoost MC) to classify dermoscopy images. Isasi et al. [9] devised a solution in the automated diagnosis landscape by curating an algorithm based on the ABCD rule and pattern recognition. Blum et al. [10] and Kiran et al. [11] proposed an automated skin lesion classification system using image segmentation, feature calculation, and classification. Fidan et al. [12] modeled a resilient approach to detecting benign and malignant skin tumors by introducing artificial neural networks (ANNs).

Esteva et al. [13] revolutionized melanocytic lesion classification by broadening horizons of deep convolutional neural networks (DCNNs) to classify skin nevi. CNNs are known to boost model performance by extracting features automatically. Treading on this path, Kawahara et al. [14] introduced a multi-resolution-tract CNN with hybrid nevus-trained layers. Brinker et al. [15] proposed an algorithm for the binary classification of melanocytic nevi with malignant nevi and compared the results obtained by the model with those of 157 dermatologists for 100 test images. Among 157 dermatologists, 136 were outperformed by CNN. Marchetti et al. [16], Haenssle et al. [17], Yu et al. [18], Maron et al. [19], Tschandl et al. [20] and Han et al. [21] performed similar analyses of CNNs vs dermatologists and proved that CNNs can fairly outperform even trained professionals.

Furthermore, skin images are employed for reliable testing and prediction by trained models to identify skin nevi images as benign or malignant. Thus, data-driven deep learning approaches may identify skin cancer correctly before they assume fatality and distinguish it from healthy tissues [22]. By utilizing multiple layers of nonlinear data processing, these methods extract, detect patterns, and define features. A deep learning model defines how to characterize images, text, or voice across multiple domains [23]. Similar to this, when a model is set up to work with large data sources, the pixels in an image are changed to internal vector highlights to categorize classes in the data. As a result, there is a dire demand to unravel the deep learning-based ensembles to classify nevus images.

The classification techniques aid in two ways: providing exact data and treatment choices for a lesion and identifying early skin cancer with adequate sensitivity and precision.

The only disadvantage to using CNNs is satisfying the data demands of this data-hungry ensemble. To overcome the demand for voluminous training data sources, transfer learning (TL) entered the limelight. TL is a machine learning method that allows users to leverage a pre-trained model as the threshold for a new classification/detection task. A model that has been previously trained on one use case with millions of images can be repurposed for a second, related use case to boost model speed and performance. Anabik et al. [24] and Pereira et al. [25] fine-tuned and leveraged multiple TL networks, including ResNet50 [26], DenseNet [27], and MobileNet [28], only to yield unsatisfactory performance metrics.

In this study, we classify skin nevi into benign and malignant types using five cutting-edge transfer learning techniques: EfficientNetB0 [29], VGG-16 [30], ResNet50 [26], InceptionV3 [31], and AlexNet [32]. We have improved the aforementioned models to deliver favorable results with reduced processing times, and we have thoroughly compared them to determine which model produces the most favorable performance metrics.

## 3 Methodology

In this section, we shall be discussing the dataset used, pre-processing methods, CNN architectures, and the evaluation metrics used.

### 3.1 Dataset and Experimental Setup

We have used the ISIC open-access database on Kaggle that contains 1800 benign and malignant melanocytic nevus (skin mole) training images. The size of each image is $224 \times 224$ pixels. For the purposes of testing, 300 and 360 malignant and benign test images were used, respectively. A validation sample size 0f 20% was taken on the training data to validate the performance of the model and obtain validation metrics. Thus, 360 validation images each were used for both the benign and malignant classes in the dataset.

We establish a regularization parameter of 0.0001 to avoid overfitting. The weight and the momentum rate of 0.9 and $e^{-4}$ are chosen as the decay parameters. The usage of pretrained weights in ImageNet returned more robust results. The network architectures are executed in Python using the Keras API with Tensorflow on an Intel(R) Core (TM) i7-10510U CPU @ 1.80GHz and 2.30GHz processors with an NVIDIA Tesla P100 GPU containing 16GB of HBM2 vRAM.

### 3.2 Pre-processing Methods

Images of melanocytic nevi must be processed to remove noise, defects, inadequate illumination, and other issues with manual imaging to make them suitable for automated diagnosis [33-34]. A few of the pre-processing techniques used to improve the image quality include boosting the overall image contrast, and color space transformations; reducing non-skin noise, and fixing illumination deficiencies such as shadows and reflections.

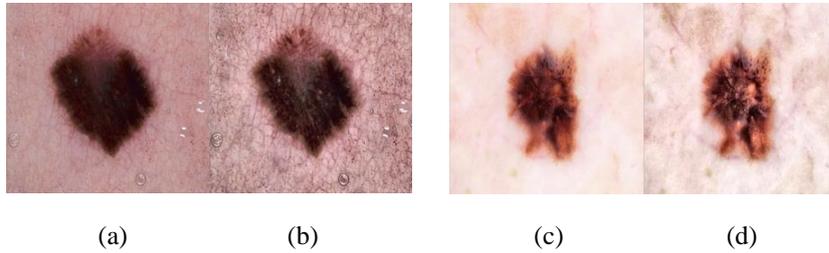

(a)     (b)     (c)     (d)

**Fig. 1.** Original Vs Pre-processed data. (a) Original dermoscopy image for benign nevus, (b) Pre-processed image, (c) Original dermoscopy image for malignant nevus, (d) Pre-processed image

### 3.3 The AlexNet Architecture

AlexNet [35] was the winner of the 2012 ILSVR challenge [36]. Proposed by Krishevsky et al., AlexNet is known to have transformed AI research. There are eight trainable layers in the AlexNet. Except for the output layer, each of the model's five layers—which combine max pooling with three fully-connected convolutional layers uses ReLu activation. Dropout layers prevented the model from overfitting. To guarantee that each neuron predicts the class probabilities of a specific image, the SoftMax activation function was used for the output. ImageNet dataset, which comprises close to 14 million images across a thousand classes, was used to train AlexNet.

### 3.4 The VGGNet Architecture

The Visual Geometry Group (VGG) introduced VGGNet in 2014. It produced the best results in the 2014 ILSVR Challenge. It also confirms that a key factor in achieving better performance is the use of much smaller filters ($3 \times 3$) to enhance the depth of the network. This architecture comes in two variations with various depths and layers. These two architectures are VGG16 and VGG19. Deeper than VGG16 is VGG19. VGG19 has more parameters and is more computationally expensive to train. For the sake of computational efficiency, the VGG16 architecture has been used. VGG16 comprises 21 layers total—13 convolutional layers, 3 dense layers, and 5 max-pooling layers—but only 16 of those layers are weight layers or learnable parameters layers.

### 3.5 The ResNet Architecture

The deep residual neural network (ResNet) architecture bagged leading places in the 2015 ILSVRC and COCO competitions. ResNet enables researchers to train neural networks that are deeper, more efficient, and more accurate in identifying objects. One Max-pool layer, one average pool layer, and 48 convolution layers make up the ResNet50 version of the ResNet model. The bottleneck features are used by ResNets at different depths to boost efficiency.

### 3.6 The Inception V3 Architecture

Inception V3 is a better version of Inception V1, a foundational model that was first released as GoogleNet in 2014. Dimensionality reduction through the decomposition of bigger convolutions into smaller, asymmetric convolutions (less than $3 \times 3$) is what gives Inception V3 its resilience. Auxiliary classifiers are used to improve the convergence of very deep neural networks. As a result, Inception V3 model architecture's auxiliary classifiers regularize classification. After increasing the activation dimension of the network filters, efficient grid size reduction is performed to lower the grid size of the feature maps. The 42 layers that make up the Inception V3 model contribute to its excellent performance.

### 3.7 The EfficientNet Architecture

EfficientNet is a DCNN scaling and design technique that consistently scales all breadth, depth, and resolution variables using a compound coefficient. In contrast to the traditional approach, which scales these variables arbitrarily, the EfficientNet scaling approach equally adjusts network width, depth, and resolution using a set of predefined scaling factors. With 237 convolutional layers, EfficientNetB0 is the most computationally efficient variant of EfficientNets. EfficientNet was also initially trained using the ImageNet dataset.

### 3.8 Evaluation Metrics

Several standard and advanced evaluation metrics have been used to analyze the efficacy of the aforementioned architectures for classifying melanocytic nevi images. The performance metrics used for comparison are accuracy, precision, recall, sensitivity, specificity, and area under the ROC (receiver operating characteristic) curve. False negatives (FN) are the number of instances that failed to accurately forecast, while true positives (TP) are the instances that did so. False positives (FP) are known as the number of negative occurrences that were wrongly predicted, whereas true negatives (TN) are the number of negative cases that were correctly predicted.

Sensitivity is the capacity of a test to identify a diseased person as positive [37]. A highly sensitive test produces minimal false negative findings.

$$Sensitivity\ or\ True\ Positive\ Rate\ (TPR) = \frac{TP}{(TP+FN)} \qquad (1)$$

Specificity is measured by its capacity to label someone as negative who doesn't have the disease [38].

$$Specificity\ or\ True\ Negative\ Rate\ (TPR) = \frac{TN}{(TP+FP)} \qquad (2)$$

Accuracy is used to show the number of correctly classified nevi divided by the total number of classified nevi. It can be represented as:

$$Accuracy\ (\%) = \frac{TP+TN}{TP+TN+FP+FN} \times 100 \qquad (3)$$

Precision is determined by dividing the genuine positives by any positive prediction [39]. It measures how well a model can predict a particular category.

$$Precision = \frac{TP}{(TP+FP)} \qquad (4)$$

Recall (also known as the True Positive Rate) is computed by dividing the real positives by anything that should have been predicted as positive [39].

$$Recall = \frac{TP}{(TP+FN)} \qquad (5)$$

The AUC-ROC curve is an indicator of performance for classification issues at different threshold levels. AUC stands for the level or measurement of separability, and ROC is a probability curve. It demonstrates how well the model can distinguish between classes [40]. AUC ROC for a binary solution is

$$AUC - ROC\ Curve = \frac{1 + TPR - FPR}{2} \qquad (6)$$

## 4 Results and Discussion

The validation metrics of the TL-based classification models AlexNet, Inception V3, VGGNet16, EfficientNetB0, ResNet50 are given in Table 1. The results show that AlexNet returns the best performance on 5 out of the 6 metrics mentioned above (accuracy, precision, recall, sensitivity, specificity). Inception V3 returns the best AUC (Area under curve) value of 0.9009, as can be seen from the plot in Fig.2. This, however, returns a deviation of merely 2.8% from AlexNet. VGG16 returns the second-best AUC-ROC value.

It is observed that AlexNet has the best accuracy of 0.9757. It is followed by VGG16 (in Fig.3.), with an accuracy of 0.8576. AlexNet and VGG16 are also the two best models in terms of precision and recall as can be observed from Fig. 4. Inception V3 returns the third-best result in the accuracy metric. It can be seen from Fig.5.

that it has the second-to-best result in the sensitivity and specificity metrics, only after AlexNet. ResNet50 and VGG16 return robust sensitivity and specificity values. ResNet50, however, has performed relatively poorly in the accuracy metric and differs from the top model (AlexNet) by 23.7%. VGG16 also has significantly good values for specificity and sensitivity, but models like AlexNet, ResNet50, and Inception V3 perform better than it does concerning these metrics. It can be visually perceived from Fig. 5. that AlexNet returned the best sensitivity and specificity values out of the 5 models. However, AlexNet was the second-most time-consuming model only after VGG16. It can be seen from Table 1 that EfficientNet performed the worst in all aspects and returned poor performance with an accuracy deviation of 43.6% from AlexNet. However, EfficientNet was the most time-efficient model to train.

All these results state that AlexNet was the most suitable TL algorithm for melanocytic nevi classification. However, it is extremely time-consuming and computationally expensive, as can be seen in Table 2. Inception V3 returned the best value of AUC ROC, while VGG16 had robust values for accuracy, precision, recall, and ROC. ResNet50 also had good results in sensitivity and specificity but lacked precision, recall, and accuracy metrics.

**Table 1.** Performance Metrics

| Model Name | Accuracy | Precision | Recall | Sensitivity | Specificity | ROC |
|---|---|---|---|---|---|---|
| AlexNet | 0.9757 | 0.9681 | 0.9795 | 1.0000 | 1.0000 | 0.8729 |
| EfficientNetB0 | 0.5396 | 0.5227 | 0.5227 | 0.5530 | 0.5530 | 0.5537 |
| VGG16 | 0.8576 | 0.8576 | 0.8576 | 0.8985 | 0.8985 | 0.8821 |
| Inception V3 | 0.8303 | 0.8131 | 0.7795 | 0.9491 | 0.9665 | 0.9009 |
| ResNet50 | 0.7386 | 0.7386 | 0.7386 | 0.9205 | 0.9205 | 0.8257 |

**Table 2.** Time Taken by Models to Train

| Model Name | AlexNet | EfficientNetB0 | Inception V3 | ResNet50 | VGG16 |
|---|---|---|---|---|---|
| Time (in seconds) | 713.21 | 158.51 | 555.38 | 267.33 | 910.92 |

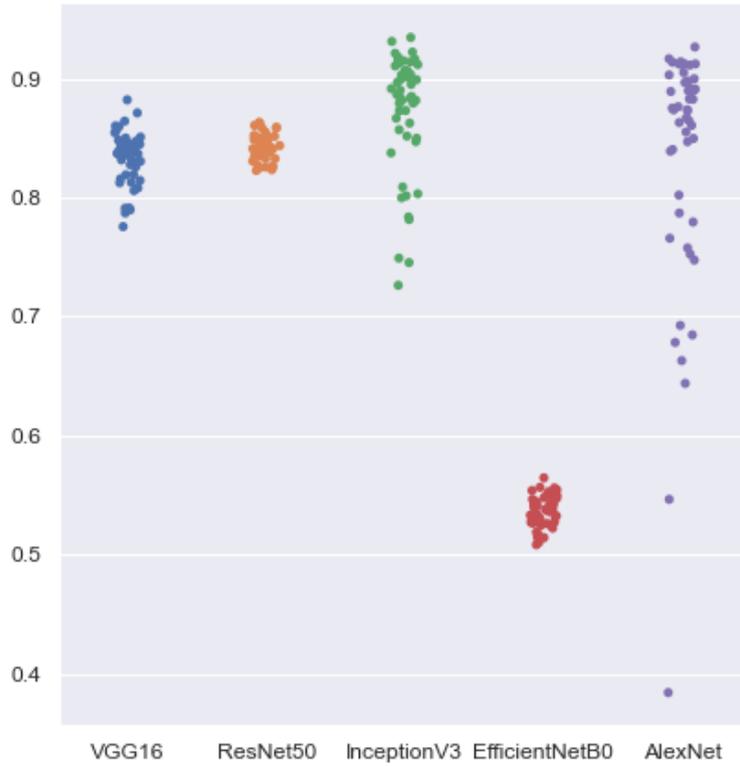

**Fig. 2.** Area under the ROC curve of the five TL architectures studied.

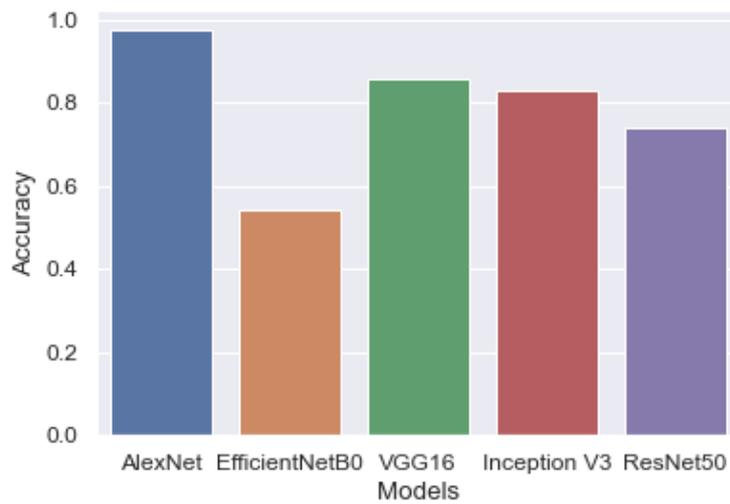

**Fig. 3.** Final Accuracy of the five transfer learning architectures studied

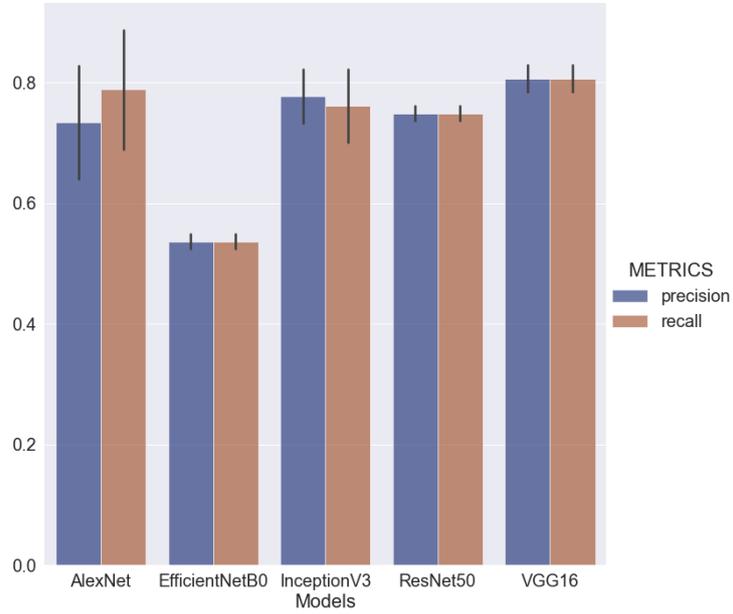

**Fig. 4.** Precision and Recall of the five TL architectures studied.

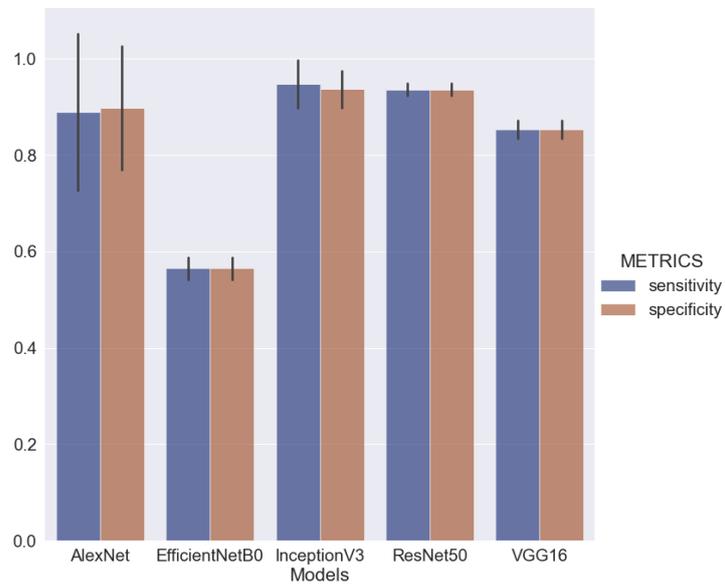

**Fig. 5.** Sensitivity and Specificity of the five TL architectures studied.

## 5      Conclusion and Future Work

To classify benign and malignant melanocytic nevi, a thorough comparative examination of five transfer-learning algorithms is conducted in this paper. The images in the dataset have undergone pre-processing to increase image clarity, contrast, and brightness. The performances of AlexNet, EfficientNetB0, Inception V3, ResNet50, and VGG16 were assessed using several common evaluation metrics, such as accuracy, precision, recall, sensitivity, and specificity. With an accuracy of up to 0.9757, our experiments demonstrated that AlexNet was the best neural network for the specified use case. It also had the highest values for the parameters of sensitivity, specificity, accuracy, and recall. However, Inception V3 turned out to be the most robust and computationally efficient DCNN. Moreover, it obtained the highest AUC-ROC score. With second-best scores for accuracy, precision, recall, and AUC-ROC, VGG16 was a potent yet the most computationally expensive model.

In the future, we intend to combine the best qualities of AlexNet and VGG16 to produce a robust and computationally efficient DCNN that will accurately classify the myriad variants of skin cancer.

## References

1. Khan M.Q., Hussain A., Rehman S.U., Khan U., Maqsood M., Mehmood K., Khan M.A. Classification of Melanoma and Nevus in Digital Images for Diagnosis of Skin Cancer. IEEE Access. 2019;7:90132–90144.
2. Popescu, D., El-Khatib, M., El-Khatib, H., & Ichim, L. (2022). New Trends in Melanoma Detection Using Neural Networks: A Systematic Review. *Sensors*, *22*(2), 496.
3. Nida, N., Shah, S. A., Ahmad, W., Faizi, M. I., & Anwar, S. M. (2022). Automatic melanoma detection and segmentation in dermoscopy images using deep RetinaNet and conditional random fields. *Multimedia Tools and Applications*, 1-21.
4. Gurung, S., and Gao, Y. (2020). "Classification of melanoma (skin cancer) using convolutional neural network," in 2020 5th International Conference on Innovative Technologies in Intelligent Systems and Industrial Applications (CITISIA) (Sydney, NSW), 1–8.
5. Pölönen, I., Rahkonen, S., Annala, L. A., and Neittaanmäki, N. (2019). "Convolutional neural networks in skin cancer detection using spatial and spectral domain," in Photonics in Dermatology and Plastic Surgery 2019 (San Francisco, CA), 302–309.
6. Anas, Mohd, Kailash Gupta, and Shafeeq Ahmad. "Skin cancer classification using K-means clustering." International Journal of Technical Research and Applications 5.1 (2017): 62-65.
7. Ebtihal A, Arfan JM (2016) Classification of Dermoscopic skin cancer images using color and hybrid texture features. Int J Comput Sci Netw Secur 16(4):135–139
8. Qaisar A, Celebi ME, Carmen S, Fondón GI, Ma G (2013) Pattern classification of dermoscopy images: a perceptually uniform model. Pattern Recogn 46:86–97
9. Isasi AG, Zapirain GB, Zorrilla MA (2011) Melanomas non-invasive diagnosis application based on the ABCD rule and pattern recognition image processing algorithms. Comput Biol Med 41:742–755

10. Blum A, Luedtke H, Ellwanger U, Schwabe R, Rassner G, Garbe C (2004) Digital image analysis for diagnosis of cutaneous melanoma. Development of a highly effective computer algorithm based on analysis of 837 melanocytic lesions. Br J Dermatol 151:1029–1038
11. Ramlakhan K, Shang Y (2011) A Mobile Automated Skin Lesion Classification System. In: Proceedings of 23rd IEEE International Conference on Tools with Artificial Intelligence, p 138–141
12. U. Fidan, I. Sari, R. K. Kumrular, "Classification of skin lesions using ann", 2016 Medical Technologies National Congress (TIPTEKNO), 2016, pp. 1–4.
13. A. Esteva et al., "Dermatologist-level classification of skin cancer with deep neural networks," Nature, vol. 542, no. 7639, pp. 115–118, Feb. 2017.
14. J. Kawahara and G. Hamarneh, "Multi-resolution-Tract CNN with Hybrid Pretrained and Skin-Lesion Trained Layers," Springer, Cham, 2016, pp. 164–171.
15. T.J. Brinker, A. Hekler, A.H. Enk, J. Klode, A. Hauschild, C. Berking, B. Schilling, S. Haferkamp, D. Schadendorf, S. Frohling, et al., A convolutional neural network trained with dermoscopic images performed on par with 145 dermatologists in a clinical melanoma image classification task, Eur. J. Canc. 111 (2019) 148–154.
16. N.C. Codella, D. Gutman, M.E. Celebi, B. Helba, M.A. Marchetti, S.W. Dusza, A. Kalloo, K. Liopyris, N. Mishra, H. Kittler, et al., Skin lesion analysis toward melanoma detection: a challenge at the 2017 international symposium on biomedical imaging (isbi), hosted by the international skin imaging collaboration (isic), in: Biomedical Imaging (ISBI 2018), 2018 IEEE 15th International Symposium on, IEEE, 2018, pp. 168–172.
17. H.A. Haenssle, C. Fink, R. Schneiderbauer, F. Toberer, T. Buhl, A. Blum, A. Kalloo, A.B.H. Hassen, L. Thomas, A. Enk, et al., Man against machine: diagnostic performance of a deep learning convolutional neural network for dermoscopic melanoma recognition in comparison to 58 dermatologists, Ann. Oncol. 29 (8) (2018) 1836–1842.
18. Y. Jiang, J. Xiong, H. Li, X. Yang, W. Yu, M. Gao, X. Zhao, Y. Ma, W. Zhang, Y. Guan, et al., Recognizing basal cell carcinoma on smartphone-captured digital histopathology images with deep neural network, Br. J. Dermatol. (2019).
19. R.C. Maron, M. Weichenthal, J.S. Utikal, A. Hekler, C. Berking, A. Hauschild, A. H. Enk, S. Haferkamp, J. Klode, D. Schadendorf, et al., Systematic outperformance of 112 dermatologists in multiclass skin cancer image classification by convolutional neural networks, Eur. J. Canc. 119 (2019) 57–65.
20. P. Tschandl, C. Rosendahl, B.N. Akay, G. Argenziano, A. Blum, R.P. Braun, H. Cabo, J.-Y. Gourhant, J. Kreusch, A. Lallas, et al., Expert-level diagnosis of nonpigmented skin cancer by combined convolutional neural networks, JAMA Dermatol. 155 (1) (2019) 58–65.
21. S.S. Han, M.S. Kim, W. Lim, G.H. Park, I. Park, S.E. Chang, Classification of the clinical images for benign and malignant cutaneous tumors using a deep learning algorithm, J. Invest. Dermatol. 138 (7) (2018) 1529–1538.
22. Bansal, P., Garg, R., & Soni, P. (2022). Detection of melanoma in dermoscopic images by integrating features extracted using handcrafted and deep learning models. *Computers & Industrial Engineering*, *168*, 108060.
23. Akib Mohi Ud Din Khanday, Bharat Bhushan, Rutvij H. Jhaveri, Qamar Rayees Khan, Roshani Raut, Syed Tanzeel Rabani, "NNPCov19: Artificial Neural Network-Based Propaganda Identification on Social Media in COVID-19 Era", *Mobile Information Systems*, vol. 2022,ArticleID 3412992, 10 pages, 2022. https://doi.org/10.1155/2022/3412992.
24. Pal, Anabik, Sounak Ray, and Utpal Garain. "Skin disease identification from dermoscopy images using deep convolutional neural networks."arXiv preprint arXiv:1807.09163 (2018).

25. F. Pereira dos Santos and M. Antonelli Ponti, "Robust Feature Spaces from Pre-Trained Deep Network Layers for Skin Lesion Classification," in 2018 31st SIBGRAPI Conference on Graphics, Patterns, and Images (SIBGRAPI), 2018, pp. 189–196.
26. He, Kaiming, Xiangyu Zhang, Shaoqing Ren, and Jian Sun. "Deep residual learning for image recognition." In Proceedings of the IEEE conference on computer vision and pattern recognition, pp. 770-778. 2016.
27. Huang, Gao, Zhuang Liu, Laurens Van Der Maaten, and Kilian Q. Weinberger."Densely connected convolutional networks." In Proceedings of the IEEE conference on computer vision and pattern recognition, pp. 4700-4708. 2017.
28. Howard, Andrew G., Menglong Zhu, Bo Chen, Dmitry Kalenichenko, Weijun Wang, Tobias Weyand, Marco Andreetto, and Hartwig Adam."Mobilenets: Efficient convolutional neural networks for mobile vision applications." arXiv preprint arXiv:1704.04861 (2017).
29. Karar Ali, Zaffar Ahmed Shaikh, Abdullah Ayub Khan, Asif Ali Laghari, Multiclass skin cancer classification using EfficientNets – a first step towards preventing skin cancer, Neuroscience Informatics,Volume 2, Issue 4, 2022.
30. Y. Jusman, I. M. Firdiantika, D. A. Dharmawan and K. Purwanto, "Performance of Multi Layer Perceptron and Deep Neural Networks in Skin Cancer Classification," 2021 IEEE 3rd Global Conference on Life Sciences and Technologies (LifeTech), 2021, pp. 534-538.
31. A. Demir, F. Yilmaz and O. Kose, "Early detection of skin cancer using deep learning architectures: resnet-101 and inception-v3," 2019 Medical Technologies Congress (TIPTEKNO), 2019, pp. 1-4.
32. Hosny KM, Kassem MA, Foaud MM. Classification of skin lesions using transfer learning and augmentation with Alex-net. PLoS One. 2019 May 21;14(5):e0217293.
33. Joseph, S., & Olugbara, O. O. (2022). Preprocessing effects on performance of skin lesion saliency segmentation. *Diagnostics*, *12*(2), 344.
34. Tang, T., Jiao, D., Chen, T., & Gui, G. (2022). Medium-and Long-Term Precipitation Forecasting Method Based on Data Augmentation and Machine Learning Algorithms. *IEEE Journal of Selected Topics in Applied Earth Observations and Remote Sensing*, *15*, 1000-1011.
35. Krizhevsky, Alex & Sutskever, Ilya & Hinton, Geoffrey. (2012). ImageNet Classification with Deep Convolutional Neural Networks. Neural Information Processing Systems. 25. 10.1145/3065386.
36. Deng, J., Dong, W., Socher, R., Li, L.-J., Li, K., & Fei-Fei, L. (2009). Imagenet: A large-scale hierarchical image database. In *2009 IEEE conference on computer vision and pattern recognition* (pp. 248–255).
37. Hill, A., & Nantel, J. (2022). Sensitivity of discrete symmetry metrics: Implications for metric choice. *PloS one*, *17*(5), e0268581.
38. Whitney, H. M., Drukker, K., & Giger, M. L. (2022). Performance metric curve analysis framework to assess impact of the decision variable threshold, disease prevalence, and dataset variability in two-class classification. Journal of Medical Imaging, 9(3), 035502.
39. Su, F., Zhang, Y., Li, F., & Ji, D. (2022). Balancing Precision and Recall for Neural Biomedical Event Extraction. *IEEE/ACM Transactions on Audio, Speech, and Language Processing*, *30*, 1637-1649.
40. Toft, J. H., Økland, I., & Dalen, I. (2022). ROC-curves—fundamentals for proper use. *Endocrine*, *76*(2), 505-505.